# Discerning the painter's hand: machine learning on surface topography

F. Ji[1†], M.S. McMaster[2†], S. Schwab[2†], G. Singh[2†], L.N. Smith[3†], S. Adhikari[2], M. O'Dwyer[2], F. Sayed[5], A. Ingrisano[6], D. Yoder[4], E.S. Bolman[3], I.T. Martin[2††], M. Hinczewski[2††], K.D. Singer[2††]

[1] Case Western Reserve University, Department of Biology, Cleveland, OH 44106.
[2] Case Western Reserve University, Department of Physics, Cleveland, OH 44106.
[3] Case Western Reserve University, Department of Art History and Art, Cleveland, OH 44106.
[4] Cleveland Museum of Art, Cleveland, OH 44106.
[5] Hathaway Brown School, Shaker Heights, OH 44122.
[6] Cleveland Institute of Art, Cleveland, OH 44106.

[†] These authors contributed equally to this work
[††] Senior project leads

**Abstract**

Attribution of paintings is a critical problem in art history. This study extends machine learning analysis to surface topography of painted works. A controlled study of positive attribution was designed with paintings produced by a class of art students. The paintings were scanned using a confocal optical profilometer to produce surface data. The surface data were divided into virtual patches and used to train an ensemble of convolutional neural networks (CNNs) for attribution. Over a range of patch sizes from 0.5 to 60 mm, the resulting attribution was found to be 60 to 96% accurate, and, when comparing regions of different color, was nearly twice as accurate as CNNs using color images of the paintings. Remarkably, short length scales, as small as twice a bristle diameter, were the key to reliably distinguishing among artists. These results show promise for real-world attribution, particularly in the case of workshop practice.

# INTRODUCTION

Machine learning (ML) analysis for artwork is a budding methodology aimed at advancing connoisseurship, the primary method of determining the attribution of an artwork, among other applications involving artistic style. ML was successfully applied to images of paintings for tasks including detecting forgeries [1-2], classifying digital collections [3-4], and discerning an artist's style [5-8]. While many of these studies have applied ML to high-resolution photographic images of paintings, in this report, we use ML to analyze topographic data obtained by optical profilometry. Further, advancements in the resolution, speed, and availability of such non-contact profilometric measurements are growing alongside big-data methods that can handle the large datasets produced by these measurements. In paintings, surface topography reveals unintended stylistic elements embedded in the surface of the painting that may include the deposition and drying of the paint, patterns in the brushwork, physiological factors, and other aspects of the painting's creation.

A critical aspect of assessing attribution is understanding how artists utilized workshops in the creation of works of art. [9] Many notable artists, including El Greco, Rembrandt, and Peter Paul Rubens, employed workshops, of varying sizes and structures, to meet market demands for their art. Connoisseurship has been widely successful at establishing a basis for workshop attribution in art historical studies. Practitioners of connoisseurship examine the visible stylistic elements of a composition, along with material elements, condition, and other clues about the fabrication process to build a historical understanding of an artwork's attribution. Additional information on the methods of connoisseurship in technical art history, and an overview of workshop practices is presented in the supplemental information (SI). Yet, many of the specifics concerning workshop practice remain elusive. In the case of workshops, the various artists attempt to create a complete painting with a singular style, challenging the methods of connoisseurship. Further, the challenges of such attributions create conflict when the attribution is closely tied to the apparent value of objects in the art market. Hence, there is need for unbiased and quantitative methods to lend insight into disputed attributions.

We hypothesize that significant unintended stylistic information exists in the 3D surface structure embedded by the painter during the painting process, and that this information can be captured through optical profilometry. In addition, by investigating patches at much smaller length scales than any recognizable feature of the painting, differences in the artist's intrinsic and unintended style will be revealed. By focusing on small features, we move away from intended stylistic information and also allow for comparing nearby regions of the same work of art. We further propose that a measurable unintended stylistic property that exists over small enough length-scales could be used to identify different hands in the same work of art; it could then be useful in attributing historic workshop paintings.

In this report, a controlled study of paintings commissioned from several artists that mimic certain workshop practices are interrogated for indicative stylistic information. A series of twelve paintings by four artists, and their associated topographical profiles, are subject to analysis to attribute the works and to ascertain the important properties involved in those attributions. Our use of identical materials and subject matter among the artists creates a stronger focus on individual stylistic components and reflects the properties and goals of workshop paintings. The methods and controls we employ here are suited to each individual task: namely, to isolate stylistic components, to test the efficacy of ML techniques on surface topography for proper attribution among several artists, to determine the length scales involved in the ML results, to compare with photographic ML, and to provide a basis for later studies on workshop paintings.

## RESULTS AND DISCUSSION

**Designing a controlled experiment to study the so-called painter's hand**

Historical paintings have known and unknown variables that contribute to their physical states, including the materials used, the artist or artists' technique or style, and damage and restoration that have occurred over time. Each of these could contribute to the attribution of a painting. The aim of this experiment is to explore the questions that (a) the brushstroke-produced high resolution profilometry data from a painting's surface contain stylistic information (i.e., painters leave behind a measurable "fingerprint"), and (b) the data is such that ML analysis can quantitatively distinguish among painters by the topographic data. Therefore, the experimental goal is to categorize small areas from the surface of paintings by their stylometric information, without the influence of purposeful stylistic (choice of tools or materials) or factors relating to the subject of the painting.

To ensure control over the stylistic and subjective content of the test set of paintings, we enlisted nine painting students from the Cleveland Institute of Art each to create triplicate paintings of a fixed subject, a photograph of a water lily (Fig.1). Each painting was created using the same materials (paint, canvas) and tools (paintbrushes) as described in *Materials and Methods* below. In addition, the students were instructed to treat the three versions as copies. To guide our investigation, four painting specialists (three art historians and a painting conservator) grouped the paintings by artist style using traditional connoisseurship, thus restricting our investigation to four painters whose work was most stylistically similar.

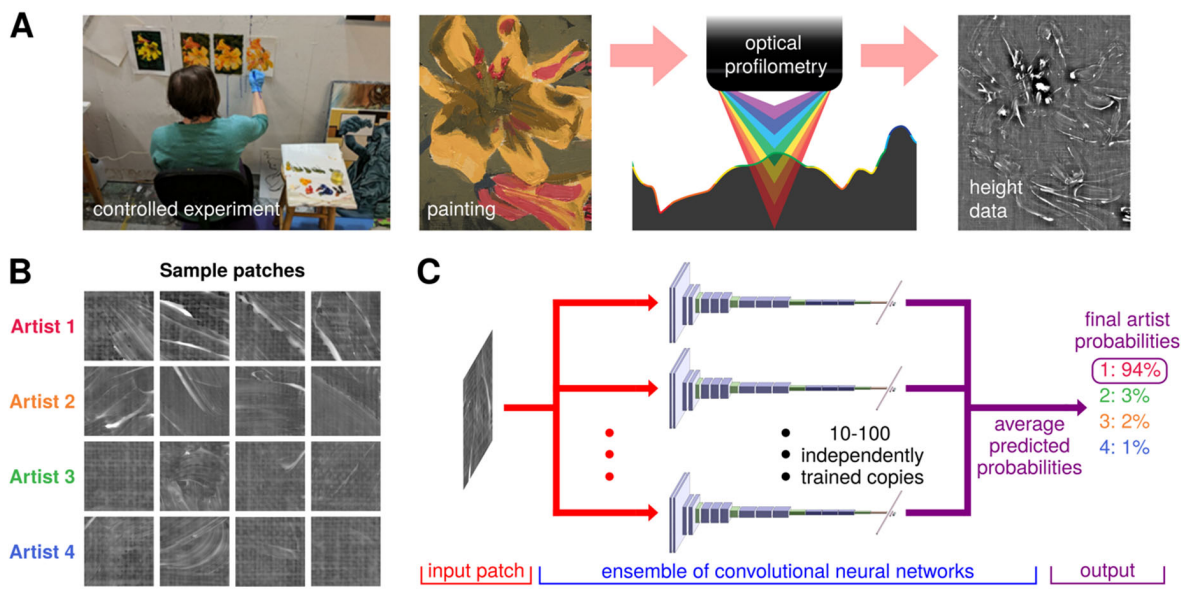

**Figure 1.** A) Overview of the data acquisition workflow. B) Sample patches of height data from each artist, each patch is 200 x 200 pixels (10 x10 mm). C) Schematic of the ML approach, which uses an ensemble of CNN to assign artist attribution probabilities to each patch.

**Acquiring and preparing data from paintings**

The surface height information for each painting was collected by high resolution optical profilometry. Measurements were conducted over a 12 x 15 cm region centered around the subject of the painting, with a spatial resolution of 50 microns, and a height repeatability of 200 nm. Given that brush strokes and their associated features are on the scale of hundreds of microns this was sufficient for capturing the fine brushstroke features of the painting's surface.

In preparation for the experiments, the height information is digitally split into small patches, the central objects of the investigation, as depicted in Fig. 1B. A typical patch size for these experiments is 1 x 1 cm, or 200 x 200 pixels, though we eventually explored a range of patch sizes from 10 pixels (0.5 mm) to 1200 pixels (6 cm). The effect of this is threefold. First, it eliminates the subjective information (the water lily figure) from the patches, since these patches are too small to individually contain indicatory information. Second, it allows us to employ ML methods for each painter. For example, the 3 paintings will provide 540 patches at the 1 x 1 cm patch size. Finally, we use ML to quantitatively attribute the individual patches of the painting. So, by reconstructing the original topography from the patches we can visually represent regions of the surface with different quantitative attributions. This will be important for our future studies inspecting historical paintings, where different regions may represent the contributions of different hands, whether from different members of the workshop or later conservation attempts.

ML methods were then applied to the surface topographic information from the paintings to explore the following questions:

1. Is there enough information to differentiate between artists?

2. Which length scales contribute useful information?

3. How does topographical information compare to photographic data?

4. What machine learning methods provide the most accurate performance?

**Training CNNs to reliably attribute patches of topographical data to individual artists**

Convolutional neural networks (CNNs) are a powerful and well-established method in computer vision tasks such as image classification. [10-11] They generally consist of three classes of layers: convolution, pooling, and fully connected layers [12] (see SI Fig. S2). Convolution layers learn translation-invariant features from the data and pooling summarizes the learned features. The stacking of these layers helps to build a hierarchical representation of the data. Fully connected layers input these extracted features into a classifier and output image classes or labels. Identification using CNNs are ideal for signals—such as topographical data—that have local spatial correlations and translational invariance. However, training a deep CNN from scratch on a small dataset typically results in a problem known as over-fitting, where the network performs better on the training set, but often does not generalize well to unseen data. To avoid this, a common solution is transfer learning [13]: adapting a network that has been pretrained on a large dataset to a different but related task. For the case of CNNs pretrained on images, the initial layers perform general feature extraction, and hence are often applicable to a broad variety of image classification problems. The final fully connected layer (and sometimes several of its predecessors) is replaced and retrained for the problem of interest. This retraining of the network is fine-tuned in a block-wise manner, starting with tuning the last few layers, and then allowing further preceding layers to be trainable as well. In this work, the network we have used is an architecture called VGG-16 [14], which was pretrained on more than one million images in the ImageNet dataset [15].

This transfer learning procedure allows us the full functionality of a highly tuned deep CNN with specificity to our task of surface topography. In short, this CNN now is outfitted to take a small input patch of the painting and produce an output pertaining to attribution. The output of our network is a 4-D vector whose components correspond to the probability of attribution to one of the four artists in the experiment (Fig. 1C). Patches from two of the three paintings from each artist are used for training / validation, with patches from the third painting reserved for testing. Because of the stochastic nature of the training procedure, involving presenting random minibatches from the training set over many epochs, the weights in the final network after training would be different if we were to repeat the whole procedure from the start. We take advantage of this stochasticity by creating ensembles of 10-100 different trained networks for each task we consider, using the mean of the probability vectors from the entire ensemble as the final prediction. We then calculate the overall accuracy as well as F1 scores, which is a measure of test accuracy for each artist. Such ensemble learning predictions in many cases outperform those of single networks [16]. Additional details of the network architecture, training, and fine-tuning procedure can be found in the *Material and Methods* and SI.

The results of ensemble ML of 100 different trained networks for attribution using patches of side-length 200 pixels (10 mm) are shown in Fig. 2. Each patch is color coded according to the largest probability (most likely artist), with the opacity of the shading proportional to the magnitude of that probability (i.e., more transparent shadings correspond to more uncertain attributions). Out of 180 patches for each artist in the test painting, we found 12, 0, 2, and 14 patches attributed incorrectly for artists 1 through 4, representing an overall accuracy of 96.1%.

This is remarkable as 25% is expected by random choice. Further, we find that most of the patches were attributed with high confidence (more opaque shading) for all four artists. The accuracy of ML prediction from the height data is remarkable, particularly given the similarity of the patches in terms of features distinguishable to the human eye (Fig. 1B) as well as its success in broad monochrome areas of the painted background.

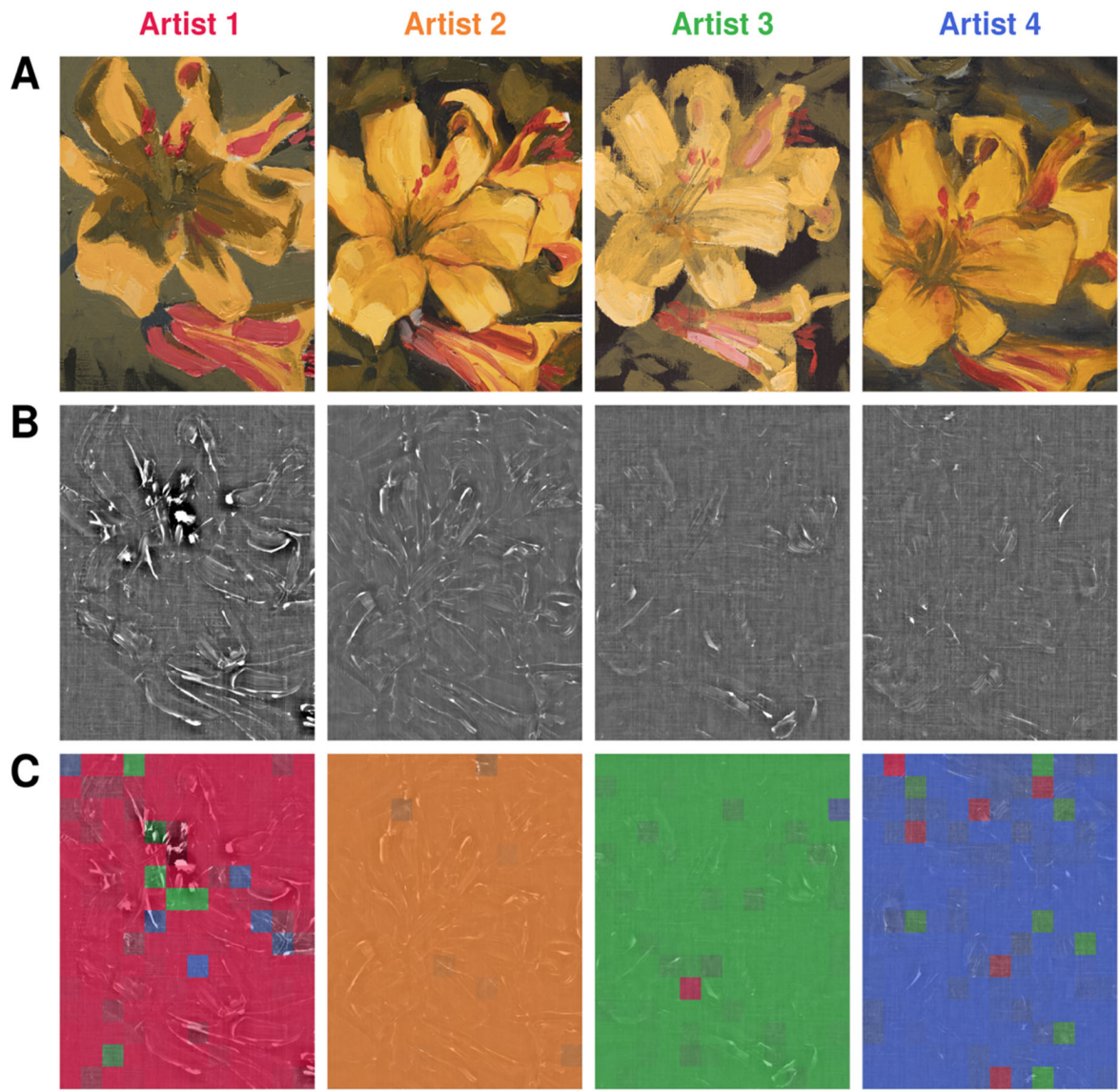

**Figure 2.** Each artist (artist 1-4) created three paintings, one of which was reserved for testing the trained ML algorithm. These test paintings are shown for all four artists as A) high-resolution photographs (all paintings used in the study are presented in SI Fig. S1), B) height data, shaded in grayscale from low (darker) to high (lighter). C) Attribution results of the ensemble ML predictions on height data. Color shadings are overlayed on the grayscale image from row B corresponding to the most likely artist for each patch of side-length 200 pixels (10 mm; 1: red, 2: orange, 3: green, 4: blue). More opaque shades of color indicate greater predictive confidence (larger attribution probabilities). The overall accuracy of the patch attribution is 96.1%.

**Exploring the effect of patch size on attribution accuracy**

The surprisingly accurate attribution of 1 cm patches leads to a natural question: how does the size of the patch affect the machine's ability to properly attribute? In other words, can we make the patch size smaller than 1 cm and still reliably attribute the hand? Fig. 3 presents results for networks trained patches with different side-lengths ranging from 10 pixels (0.5 mm) to 1200 pixels (6 cm). The predictions are quantified in terms of overall accuracy for all four artists (solid thick curve) and individual artist $F_1$ score (thin colored curves). We also calculated precision and recall, results are shown in SI Fig. S3. To check the self-consistency of the predictions, we conducted repeat training / testing trials at each patch size (details in the SI). The data points and error bars in Fig. 3 represent the mean and standard deviation for those trials.

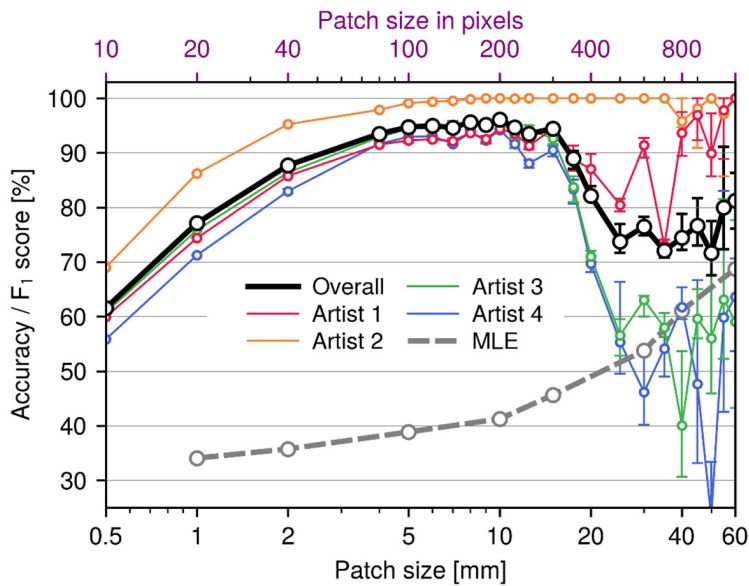

**Figure 3.** Patch size versus two measures of ML performance: overall accuracy (thick black curve) and $F_1$ scores for each artist (thin color curves). The best results occur for patches in the range 100-300 pixels. The overall accuracy decreases when patch size is smaller than 100 pixels (due to lack of information in each patch) and when patch size is larger than 300 pixels (because the size of the training data set decreases with increasing patch size). Error bars are standard deviations over different repetitions of ensemble training. For comparison, the maximum likelihood estimation (MLE) accuracy results based on the pixel height distributions are shown as a dashed curve.

The accuracy exhibits a broad plateau around 95% between 100 and 300 pixels, the optimal patch size range for attribution among these artists. Below 100 pixels there is a gradual drop-off in accuracy, as each individual patch contains fewer of the distinctive features that facilitate attribution. The $F_1$ scores allow us to separate out the network performance for each artist. Consistent with the results in Fig. 2, the attribution is generally better for artists 2 and 3 versus 1 and 4 across patch sizes less than 300. Nonetheless, the $F_1$ scores for all artists are above 90% near the optimal patch size (around 200 pixels).

On the other end of the patch size spectrum, the ML approach faces a different challenge. The size of training sets becomes quite small, even though each individual patch contains many informative features. The single-network accuracy drops off quickly for patch sizes above 300 pixels, decreasing to about 75% at the largest sizes.

**Predictions using single-pixel information versus spatial correlations**

One of the hallmarks of CNNs is their ability to harness spatial correlations at various scales in an input image in order to make a prediction. However, there is also information present at the single pixel level since each artist's height data will have a characteristic distribution relative to the mean. The probability densities for these distributions are shown in Fig. 4, calculated from the two paintings in the training sets of each artist. The height distributions are all single-peaked and similar in width, except for Artist 1, who exhibits a broader tail at heights below the mean than the others. In order to determine how important spatial correlations are, we can compare the CNN results to an alternative attribution method that is blind to the correlations: maximum likelihood estimation (MLE). For a given patch in the testing set, we calculate the total likelihood for the height values of every pixel in the patch belonging to each of the four distributions in Fig. 4. Attribution of the patch is assigned to the artist with the highest likelihood. The predictive accuracy of the MLE approach versus patch size is shown as a dashed line in Fig. 3. We expect MLE to perform the best at the largest patch sizes, since each patch then gives a larger sampling of the height distribution, and hence is easier to assign. Indeed, at the patch size of 1200 pixels, representing nearly a fifth of the area of a single painting, the MLE accuracy approaches 70%, comparable to the CNN accuracy. In this limit the size of the training set is likely too small for the CNN to effectively learn correlation features. As the patch size decreases, the gap between the CNN and MLE performance grows dramatically. In the range of 100 – 300 pixels where the CNN performs optimally (~ 95%), the MLE accuracy is only around 40%. These small patches are an insufficient sample of the distribution to make accurate attributions based on single pixel height data alone. Clearly the CNN is taking advantage of spatial correlations in the surface heights. This leads to a natural next question: what correlation length scales are involved in the attribution decision?

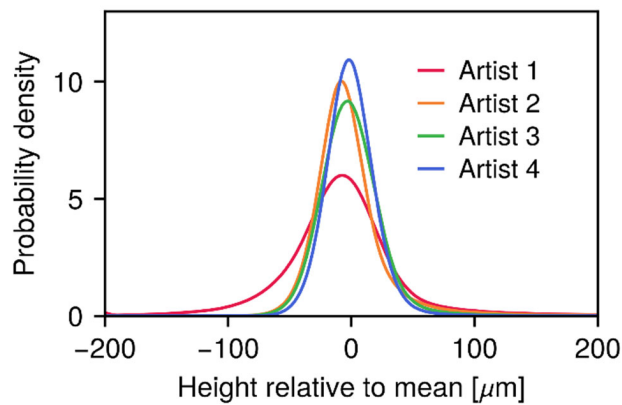

**Figure 4.** Distributions of heights relative to the mean for each artist.

**Using empirical mode decomposition to determine the length-scales of the brushstroke topography**

In order to examine the spatial frequency (length) scales most important in the ML analysis, we employed a preprocessing technique used historically in time-series signal analysis called empirical mode decomposition (EMD) [17] which has recently been extended into the spatial domain [18-20]. Its versatility is derived from its data-driven methodology, relying on unbiased techniques for filtering data into intrinsic mode functions (IMFs) that characterize the signal's innate frequency composition [21]. In our case, we have used a bi-directional multivariate EMD [22] to split our 3D reconstruction of each painting's complex surface structure into IMFs that characterize the various spatial scales present.

The first IMF contains the smallest length scale textures, and subsequent IMFs contain larger and larger features until the sifting procedure is halted and a residual is all that remains. This process is lossless in the sense that by adding each IMF and the resulting residual together, the entire signal is preserved [17], [21]. It is also unbiased in the sense that when compared to standard Fourier analysis techniques, there are no spatial frequency boundaries to define, and no edge effects introduced from defining those boundaries.

By investigating each series of IMFs individually, we can estimate the length scale for each as follows. We use a standard 2D fast-Fourier transform on the IMF and calculate a weighted average frequency for the modes. The length scale is the inverse of the average frequency, and is plotted versus IMF number in Fig. 5B. Among the four artists, the typical scale increases from about 0.2 mm for IMF 1 to 0.8 mm for IMF 5. Fig. 5A shows a sample patch and the corresponding IMFs, which illustrates the progressive coarsening at higher IMF numbers. To see how the length scale affects the attribution results, we repeated the CNN training using each IMF separately, rather than the height data. The resulting mean accuracies versus IMF number at three different patch sizes are shown in Fig. 5C. Individual IMFs are by construction less informative than the full height data (which is a sum of all the IMFs), and hence we do not reach the 95% level of accuracy seen in the earlier CNN results. However, IMFs 1 and 2 (the smallest length scales) achieve accuracies of above 80% at patch size 10 mm (200 pixels). There is a drop-off in accuracy as we go to larger patch sizes (IMFs 3-5), indicating that the salient attribution information is at lengths scales of 0.2 - 0.4 mm. These are comparable to the dimensions of a single bristle in the two types of brushes used by the artists (0.25 and 0.65 mm respectively, shown as dashed lines in Fig. 5B). This strongly suggests that the key to attribution using height data lies at scales that are small enough that they reflect the unintended (physiological) style of the artist. This result is consistent with the scale-dependent ML results depicted Fig. 3. Below a patch size of 5 mm, all results are well-above that expected for random attribution (25%). Remarkably, even at the scale of 0.5 mm, that is, the scale of 1-2 bristle widths, ML was able to attribute to 60% accuracy

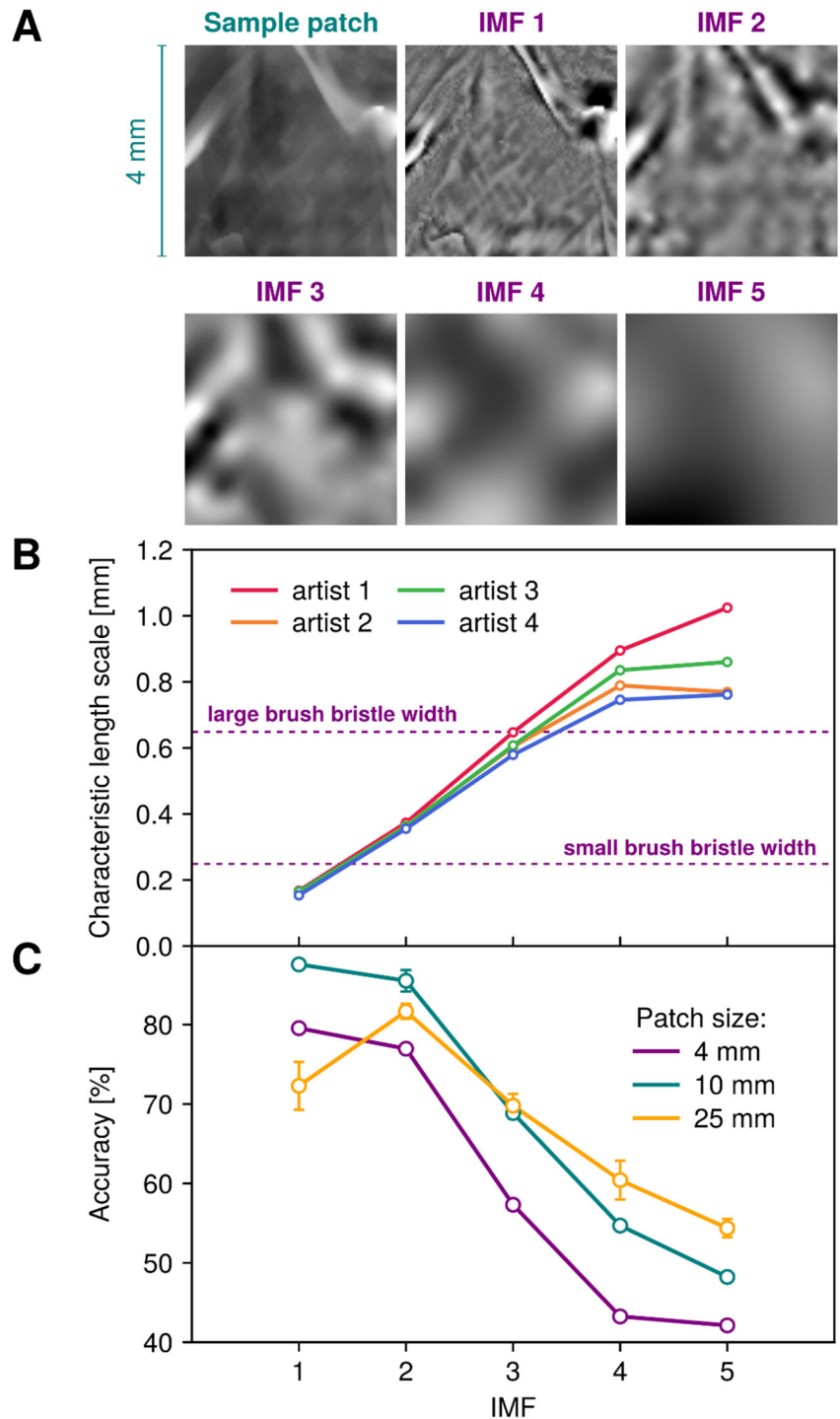

**Figure 5.** A) A sample patch of side-length 80 pixels (4 mm) and the corresponding first five IMFs calculated using empirical mode decomposition. B) The characteristic length scale of the features in each IMF for each artist. C) Mean accuracy of attribution when the network is trained on each IMF alone, rather than the original height data. Results are shown for three different patch sizes: 80 pixels (4mm), 200 pixels (10mm), and 500 pixels (25mm).

**Comparing topography versus photography when testing on data with novel characteristics**

Image recognition by ML is most often performed on photographic images of the subject depicted by arrays of RGB channels performed on the entire image. We sought to determine how well using CNNs on patches of the images of row A of Fig. 2 would perform compared to the profilometry data. We were particularly interested in how well ML of the two types of data—photo and height-based—would perform if the testing set had novel colors and subject matter absent in the training set. This approach better approximates the challenges of real-world attribution, where we would not necessarily have extensive well-attributed training data matching the palette and content of the regions of interest in a painting where the algorithm would be applied. To generate qualitatively distinct training and testing sets, we divided each painting into patches of side- length 100 pixels (5 mm) and then sorted the patches into three categories: background, foreground, and border depending on the color composition of each patch (see Fig. 6A for an example). Among our training set, 25% of the patches are assigned to background, 50% count as foreground, and the remaining 25% are border patches (Fig. 6B). The latter include regions of both background and foreground, and were excluded from both training and testing to make the analysis more stringent. The mostly dark green and black color palette and lack of defined subjects in the background distinguishes it from the foreground, which is dominated by the painted flower, with various shades of yellows and reds. Could a network trained on only background patches still accurately attribute foreground patches, or vice versa? The mean accuracy results are shown Fig. 6C, with the left two bars corresponding to training on the background, testing on the foreground, and the right two bars to the reverse scenario. Because the training sets are significantly smaller (and less representative of the test sets) than in our earlier analysis, we expect lower attribution accuracies. Despite this, networks trained on the height data (blue bars) perform reasonably well, achieving 60% accuracy when trained on background, and 80% when trained on foreground. (We note that the background training set is about half the size of the foreground one.) In contrast, networks trained on the photo data did significantly worse (red bars), achieving 27% and 43% accuracies, respectively. Clearly, in this context the color and subject information in the photo data, which was likely the focus of the ML training, was a hindrance, since the test set confronted the network with novel colors and subject matter. On the other hand, there is a significant, small-scale, stylistic component in the height data that is present whether the artist is painting the foreground or background, which can therefore be harnessed for attribution.

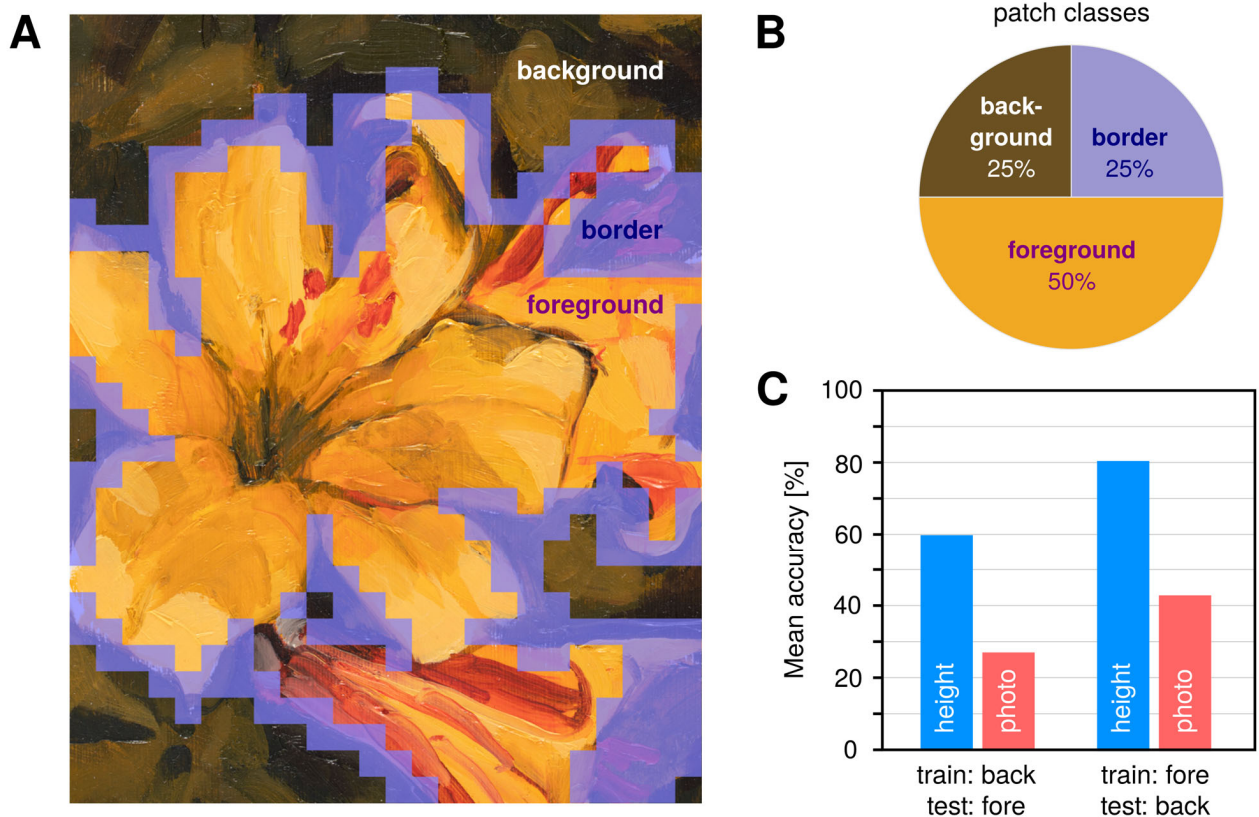

**Figure 6.** A) Example of a painting divided into background, foreground, and border (mixed background and foreground) patches of size 100 x 100 pixels (5 x 5 mm). B) Distribution of the three classes across all patches in the training set. C) Results from training on background patches, testing on foreground patches and vice versa using both height data and high-resolution photo data.

# Conclusions

We have described a controlled experiment using ML methods coupled with optical profilometry data to attribute painted works of art based on the topographical structure indicative of the artists' style as a tool for art connoisseurship. By dividing the paintings into patches significantly smaller than the painting, we have removed subject information as well as aspects of the artist's intended style. We are thereby able to focus our study onto attribution using only unintended style components. We found outstanding attribution accuracy of over 96% using ensemble CNNs, and further, provided evidence using EMD alongside ML, that the smallest length scales characteristic of a small number bristles are most telling for the ML attribution process. Additionally, we found that profilometry data provides higher attribution accuracy than using photographs when the subject and color palettes of the training and testing data are significantly different. Our virtual patch analysis is appropriate for attributing workshop paintings. In our planned application to real-world paintings, however, the toll of intervening years and conservation measures will challenge our techniques.

# MATERIALS AND METHODS

**Experimental Design**

This study was designed to assess the potential usefulness of surface topography in the attribution of paintings. Sets of triplicate paintings by a single artist were created by nine students, all using the same tools, materials, and subject. The surface topographic information for each painting was recorded using optical profilometry. Then, machine learning technology was developed to attribute small square patches of the painting's topography to the artist who painted them. Lastly, data filtering software was used to separate the height data into intrinsic mode functions (IMFs), based on spatial frequency, to determine the length scales of interest. In this way, we demonstrate the usefulness of topographic information to attributing paintings, and further investigate the method.

**Materials**:

The paintings were prepared using Winsor & Newton Winton Oil Colors: Titanium White, Cadmium Red, Cadmium Yellow, French Ultramarine, and Burnt Umber (Blick), Utrecht Linseed Oil (Blick), on cut canvas paper from a Canson Foundation Canva-Paper Pad (Blick). The paint was applied with a classroom set of off-the-shelf, Blick Scholastic Wonder White paintbrushes (sizes: Bright 8, Bright 4, Round 4), which were used at the artist's discretion.

**Preparation of the paintings**

In preparation for the OP measurements, the paintings were adhered to plexiglass donated by the Cleveland Museum of Art, using 3M Super 77 Multipurpose Spray Adhesive. This was done to ensure the entire 12 x 15 cm region of the painting stays within the 1.1 cm focal range of the optical profilometer for the duration of the measurement.

**Profilometry**: Surface profiles (12 x 15 cm) were collected using a Nanovea ST400 surface profiler based on chromatic confocal detection over an area of 12 by 15 cm with 50 um spatial resolution. Specifications for the P5 pen used for the measurement include a lateral resolution of 11 μm, a 10 mm z-range, and a height repeatability of 200 nm within that range. Three thousand 12 mm line scans were made across the samples, separated by 50 μm in the y-direction. Measurements were aligned to the center of the painting, so a favorable ratio of subject and background could be collected.

**Data preparation**

As an initial preprocessing step to remove large scale height variations due to the curvature of the canvas, we estimated the canvas profile by applying a mean filter of radius 100 pixels to the height data. The resulting profile is subtracted from the raw height data. These relative height values are then mapped so that the range from –200 to 300 microns corresponds to the grayscale range 0 to 1. The values are copied three times to make three identical channels in a 16-bit PNG image file, to meet the three-channel requirement of VGG-16. When we use CNN to analyze individual IMFs, we also copy each individual IMF three times.

Then we divide each painting into patches of various sizes ranging from 0.5mm (10 pixels) to 60mm (1200 pixels). We preprocess each patch by subtracting the mean RGB value [103.939, 116.779, 123.68], which is computed on all the training images in ImageNet database, from each pixel. The default input size for VGG-16 is 224 pixels × 224 pixels, we then resize all the patches to match this size.

Each artist created three paintings, we choose one from each artist as the test painting, and the rest two as training/validation paintings and the ratio of the number of training and validation patches is 9:1. To examine how this choice of test versus training/validation paintings affects the network performance, we also looked at variations: out of the 81 possible combinations, we randomly selected 10 and show the results in SI Fig. S4. The performance is comparable to the main text results.

**CNN architecture**

To perform transfer learning, we remove the three fully connected layers at the top of the pretrained VGG-16 network to get the base model. The convolutional base functions as a feature extractor and learns spatial hierarchies of input images. We then add a new Average Pooling layer (performed over a 2 × 2 pixel window, with stride 2), followed by a Flatten layer with 25% dropout. Next is a fully connected layer with 25% dropout, where we use ReLU activation function and apply a L2 regularization penalty with regularization factor 0.001. The final layer is a soft-max layer that performs 4-artist classification. The overall network architecture is shown in SI Fig. S2. We use Adam as the optimizer, the learning rate is set to 0.001 and 0.0001 depending on the training phase.

During training, we first randomly initialize the network weights by training the model for 25 epochs with learning rate 0.001 and batch size 32, and save the model with highest accuracy from this phase of training. Then we unfreeze the last two blocks of the VGG 16 base to fine tune the weights. During this phase of training, we use a slower learning rate 0.0001 and train the network for 25 epochs with batch size 32.

**Calculating ensemble accuracy and F1 scores**

After training, the network outputs a vector of four probabilities for each patch, corresponding to the likelihood of attribution to one of the four artists in the experiment. We create ensembles of 10-20 different trained networks depending on the patch size (more information in SI) and calculate the mean probability vector, which we use to predict which patch is painted by which artist. After making predictions of all the patches on the testing painting, we calculate the overall accuracy and F1 scores for each artist. To calculate the overall accuracy, we count the number of correctly labeled patches and divide it by the total number of testing patches. For each individual artist, we count the number of true positives (TP), false positives (FP), false negatives (FN), and calculate the F1 scores according to $F1 = TP/(TP + ½(FP + FN))$. Additionally, we also calculate precision and recall for each artist. Precision measures the proportion of patches attributed to artist X are actually from artist X, it is defined as Precision = TP/(TP + FP). Recall, on the other hand, measures the proportion of patches that are from artist X are correctly attribute to artist X, we calculate it according to Recall = TP/(TP + FN).

**Attribution based on maximum likelihood estimation (MLE)**

To understand the role of spatial correlations in attribution accuracy, we also employed an MLE approach where these correlations are ignored. To implement this method, we first did kernel density estimation (using Silverman's rule for bandwidth [23] to obtain probability density functions $P_i(z)$ for the single-pixel heights $z$ relative to the mean for each of the four artists, $i = 1$, … 4. These estimates, shown in Fig. 2, were based on combined data from the two paintings in the training set for each artist. For a given test patch, the log-likelihood that the $j$th pixel's height value $z_j$ belongs to the distribution of artist $i$ is: $\log P_i(z_j)$. The overall log-likelihood that all pixels in the patch belong to artist $i$ is: $L_i = \Sigma_j \log P_i(z_j)$, where the sum runs over all height values in the patch. MLE assigns the attribution of the patch to the artist $i$ who has the largest $L_i$ value. The resulting mean accuracies are shown as a dashed line in Fig. 5.

**Code availability**

The code associated with this manuscript is available at: https://github.com/hincz-lab/machine-learning-for-art-attribution

# ACKNOWLEDGEMENTS

We thank the following students from the Cleveland Institute of Art for creating the paintings used in this study: Emily Imka, Jace Lee, Brandon Secrest, Maeve Billings (artists 1-4 respectively), Julia O'Brien, Jamie Cohen-Kiraly, D'nae Webb, Seneca Kuchar, and Nolan Meyer. We thank the Cleveland Museum of Art Photography Department for high-resolution photographs of the paintings. We also acknowledge discussions and input by Prof. Catherine B. Scallen (Art History and Art) and Per Knutås of the Cleveland Museum of Art. This work was partially supported by a research award by Case Western Reserve University. The authors acknowledge the use of the Materials for Opto/electronics Research and Education (MORE) Center, a core facility at Case Western Reserve University (Ohio Third Frontier grant TECH 09-021). This work made use of the High Performance Computing Resource in the Core Facility for Advanced Research Computing at Case Western Reserve University for model training, testing, and deployment.

# AUTHOR CONTRIBUTIONS

AI, IM, MM, MO, FS, KS, LS, DY contributed to experimental design and data collection.
SA, MH, FJ, MO, GS, SS contributed to machine learning calculations and data presentation.
SS contributed to empirical mode decomposition
EB, LS, DY contributed to the technical art history context of the work.
All authors contributed to writing and editing the manuscript.
24

# SUPPLEMENTARY INFORMATION

**Connoisseurship, technical art history, and workshop practices**

The primary method of determining the attribution of a painting, and thus its value, is through expert connoisseurship. Connoisseurship as an art historical methodology involves the evaluation and appreciation of an object's quality, the determination of the time and place of its execution, and the identification of the artist, if possible. A key component of connoisseurship is the close study of the object in question, as the materiality of the work is critical to a proper attribution. Establishing as much as you can of the origin of an object is a crucial first step to deeper, more comprehensive scholarly study.

Western art connoisseurship has a long and richly complex history, stemming from classical antiquity and Renaissance writers on art [24]. It was not until the second half of the 19th century that a more scientific and rigorous employment of connoisseurship was developed. One of the most famous 19th century connoisseurs was Giovanni Morelli, who utilized his background in comparative anatomy and medicine to adapt principles of classification from the natural sciences to the study of paintings. Morelli's method scrutinized seemingly minor, or 'subconscious,' painted details, such as ears and fingernails, to detect the 'hands' of the artist [25]. Morelli's most famous successor was Bernard Berenson, who built upon Morelli's method to incorporate other types of evidence, including the artist's style and influence of their geographical region [26]. While connoisseurship continued to be refined by scholars, art dealers, and collectors into the 20th century, it was still an imperfect process as it was completely reliant on the 'eye' and honesty of the expert.

Since the early 20th century, there has been an increasing utilization of characterization methods to provide new evidence related to attribution and the condition of the object [27].

Common technical methods utilized by conservation scientists and conservators include X-radiography, infrared reflectography (IRR), cross-sectional analysis, and dendrochronology [28]. Over the last several decades, numerous technical collaborations have been formed to investigate paintings attributed to notable artists such as Hieronymus Bosch [29] and Rembrandt [30]. The inclusion of ingenious new applications, in conjunction with traditional art historical methods, has changed our views and understanding of standard artistic practices, including issues of materials and techniques and details about the precise nature of collaboration. Modern technical tools have also proven proficient at assessing fakes, adding to their worth in appraising the market value of an object.

Even with the inclusion of advanced technologies, scholars continue to face numerous obstacles when assessing the details surrounding the creation of an object. A particular challenge that our collaboration focuses on is the historic use of workshops by artists [31]. The workshop enabled a master or primary artist to increase their output and offer a range of products. The resulting paintings could span the range from master-only production to nearly workshop-only with slight retouching by the named artist, and they were often valued accordingly. Due to the frequent lack of primary textual evidence, it is often difficult to determine how each workshop was structured and who was a part of these endeavors. In addition to a primary or master artist, workshops could include specialized or traveling artists, apprentices, engravers, and others. Many workshop apprentices would late become successful and influential artists in their own right, such as Anthony van Dyck, who apprenticed in the workshop of Rubens.

These limitations to assessing attribution create conflict when the attribution is closely tied to the apparent value of objects in the art market. For example, the *Salvator Mundi* attributed to Leonardo da Vinci sold at auction in 2017 for $450.3 million has yet to be publicly displayed as

planned due to the contentious debate among specialists as to whether it is an autograph da Vinci or workshop product [32]. In March 2021, a painting attributed to the 'circle' of 17[th]-century Spanish painter José de Ribera and valued at $1,800 USD was removed from an auction in Spain over the potential that it may, in fact, be a long lost, but well documented Caravaggio painting [33]. If the painting is attributed to the enigmatic early modern artist, the painting will likely sell for millions at auction.

**Details of ensemble learning**

We create ensembles of 10-100 different trained networks depending on the patch size to make final predictions. Specifically, when the patches are of side-length 10 pixels, we train 10 different networks to calculate ensemble accuracy. For patches of side-length 20 pixels, we train 20 different networks. For patches of side-length 40 pixels, we train 2 groups of 20 different networks. For patches of side-length 80 pixels, we train 4 groups of 20 different networks. And for patches of side-length 100, 120, 140, 160, 180, 200 224, 250, 300, 350, 400, 500, 600, 700, 800, 900, 1000, 1100, and 1200 pixels, we train 5 groups of 20 different networks to get 5 ensemble accuracies. Errors, when present, are calculated using different ensemble accuracies. Results in Fig. 3C are predicted by an ensemble of 100 different trained networks.

# Supplementary figures

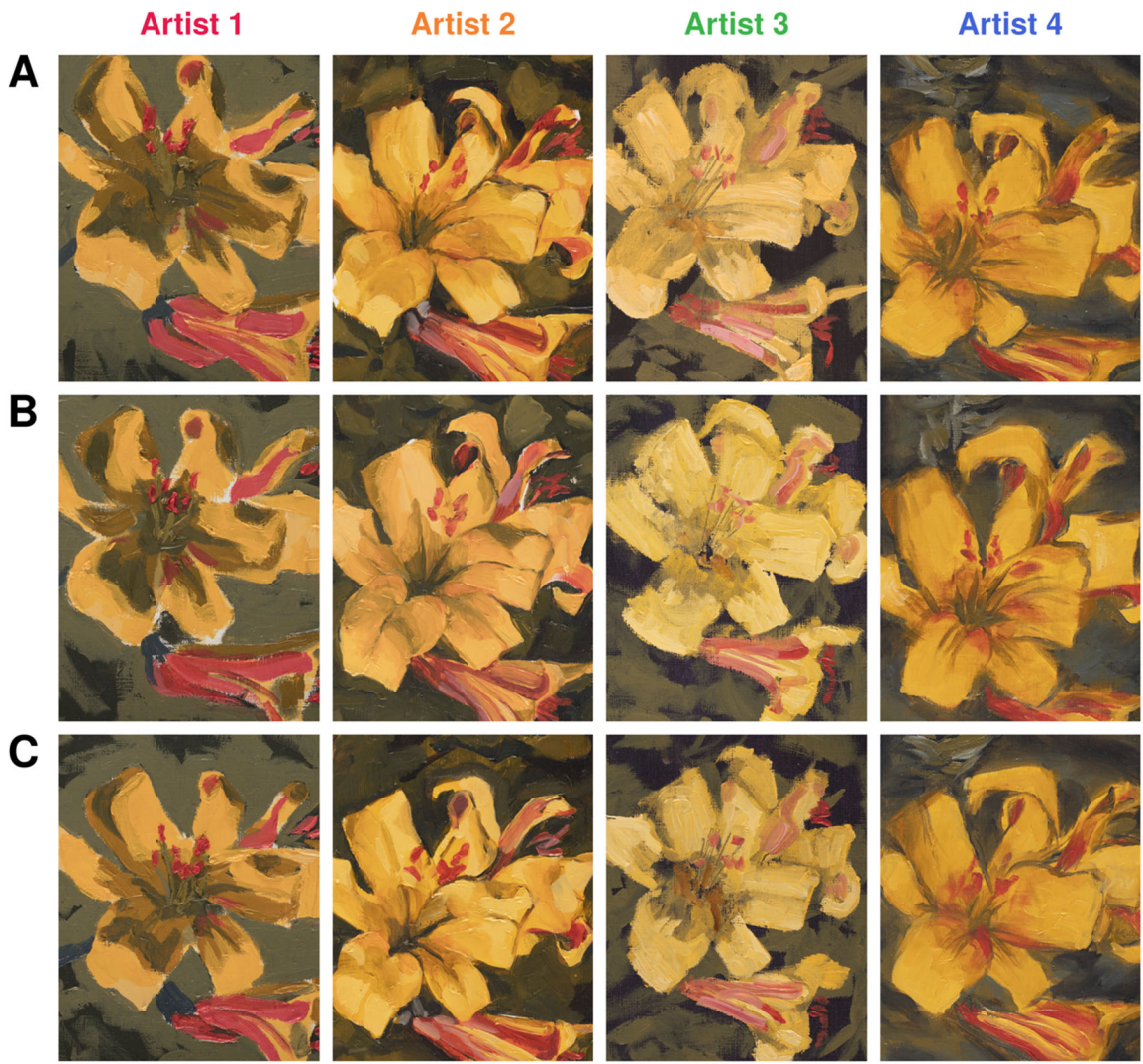

**Figure S1.** A) Test paintings from each artist. B-C) Training paintings from each artist.

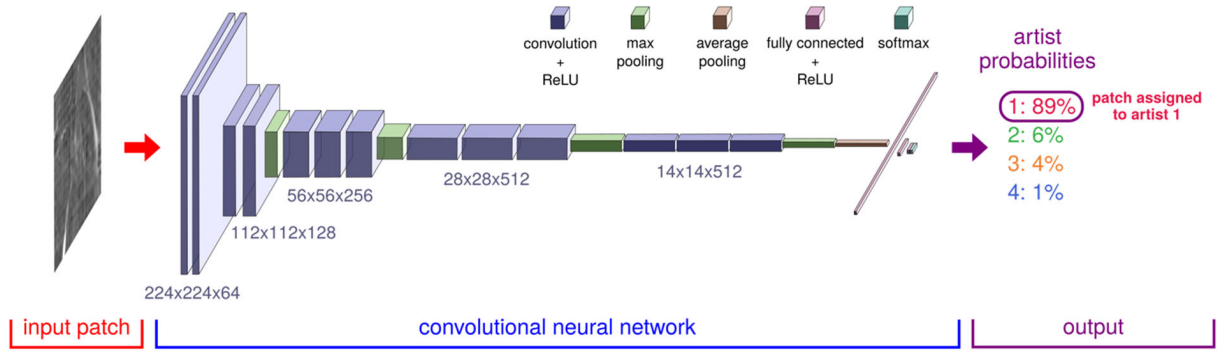

**Figure S2.** Schematic of the convolutional neural network architecture.

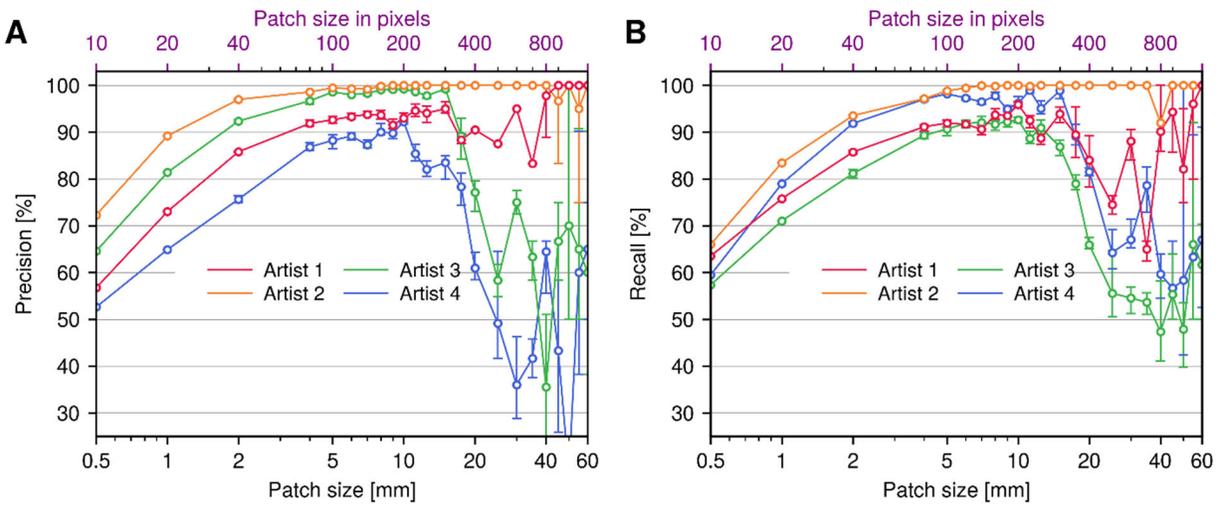

**Figure S3**. Patch size versus A) precision and B) recall for each artist.

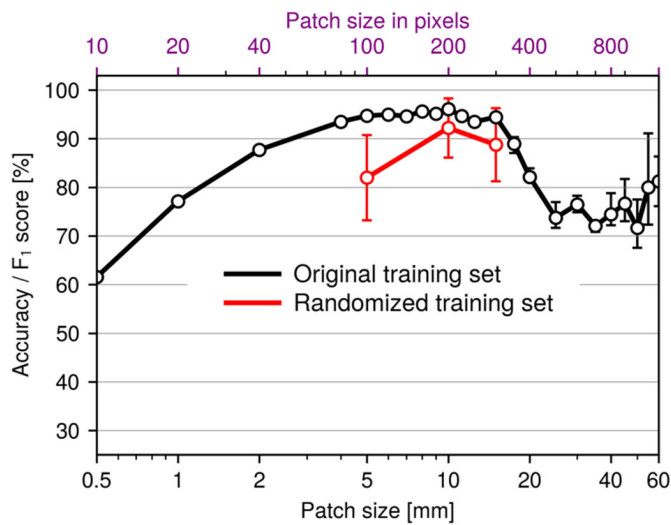

**Figure S4.** Mean accuracy versus patch size for the original choice of training set used in the main text (black curve), versus the mean results from a random selection of 10 alternative training sets (red curve), corresponding to different choices of which two out of three paintings from each artist to use for training.